\documentclass[11pt,a4paper]{article}
\usepackage{authblk}
\usepackage[hyperref]{acl2021}
\usepackage{times}
\usepackage{latexsym}

\usepackage{tablefootnote}

\usepackage{microtype}
\usepackage{multirow}
\usepackage{graphicx}
\usepackage{booktabs}
\usepackage{amsmath}
\usepackage{amsfonts}
\usepackage{float}

\aclfinalcopy 
\def\aclpaperid{599} 

\title{UniKeyphrase: A Unified Extraction and Generation Framework for Keyphrase Prediction}

\author{\textbf{Huanqin Wu}$^{1} \thanks{\quad Equal contribution.}$,
\textbf{Wei Liu}$^{1*}$,
\textbf{Lei Li}$^{2}$,
\textbf{Dan Nie}$^{1}$,
\textbf{Tao Chen}$^{1}$,
\textbf{Feng Zhang} $^{1}$,
\textbf{Di Wang} $^{1}$
 \\
$^{1}$Tencent AI Platform Department, China \\
$^{2}$Beijing University of Posts and Telecommunications \\
\tt
\normalsize{\{huanqinwu,thinkweeliu,kathynie,vitochen,jayzhang,diwang\}@tencent.com} \\ \tt
\normalsize{leili@bupt.edu.cn}}

\date{}

\begin{document}
\maketitle
\begin{abstract}
Keyphrase Prediction (KP) task aims at predicting several keyphrases that can summarize the main idea of the given document.
Mainstream KP methods can be categorized into purely generative approaches and integrated models with extraction and generation.
However, these methods either ignore the diversity among keyphrases or only weakly capture the relation across tasks implicitly.
In this paper, we propose \textbf{UniKeyphrase}, a novel end-to-end learning framework that jointly learns to extract and generate keyphrases. 
In UniKeyphrase, stacked relation layer and bag-of-words constraint are proposed to fully exploit the latent semantic relation between extraction and generation in the view of model structure and training process, respectively. 
Experiments on KP benchmarks demonstrate that our joint approach outperforms mainstream methods by a large margin.
\end{abstract}

\section{Introduction}
Keyphrases are several phrases that highlight core topics or information of a document.
Given a document, the KP task focuses on automatically obtaining a set of keyphrases. 
As a basic NLP task, keyphrase prediction is useful for numerous downstream NLP tasks such as summarization \cite{wang2013domain, pasunuru2018multi}, document clustering \cite{hulth2006study}, information retrieval \cite{kim2013applying}.

\begin{figure}[!t]
\centering\includegraphics[width=0.48\textwidth]{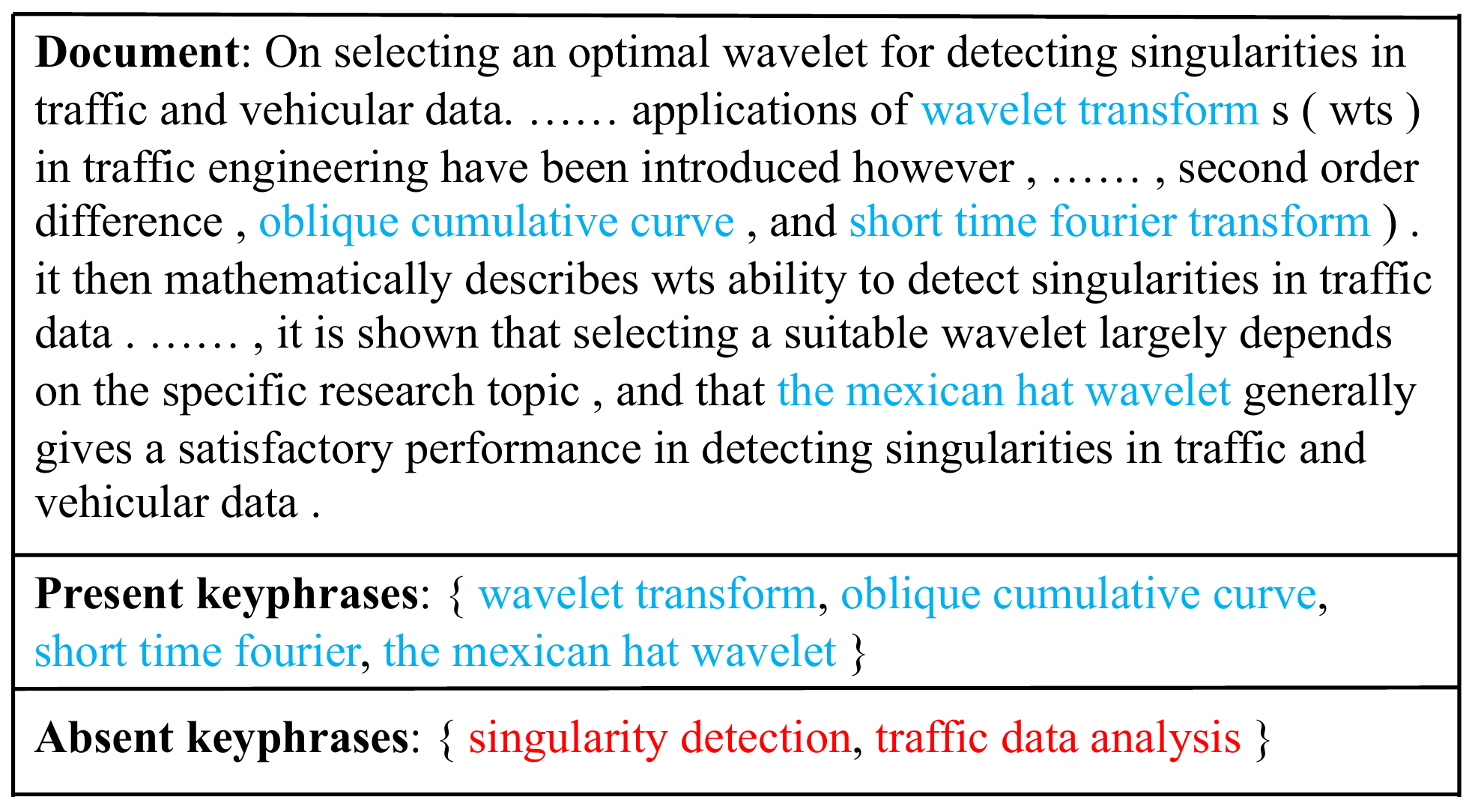}
\caption{An example of an input document and its expected keyphrases. Blue and red denote present and absent keyphrases, respectively.}
\label{fig1:case-study}
\end{figure}

Keyphrases of a document fall into two categories: \textit{present keyphrase} that appears continuously in the document, and \textit{absent keyphrase} which does not exist in the document. Figure \ref{fig1:case-study} shows an example of a document and its keyphrases.
Traditional KP methods are mainly extractive, which have been extensively researched in past decades \cite{witten2005kea, nguyen2007keyphrase, medelyan2009human, lopez2010humb, zhang2016keyphrase, alzaidy2019bi, sun2020joint}.
These methods aim to select text spans or phrases directly in the document, which show promising results on present keyphrase prediction. However, extractive methods cannot handle the absent keyphrase, which is also significant and requires a comprehensive understanding of document.

To mitigate this issue, several generative methods \cite{meng2017deep, chen2018keyphrase, ye2018semi, wang2019topic, chen2019guided, chan2019neural, zhao2019incorporating, chen2020exclusive, yuan2020one} have been proposed. Generative methods mainly adopt the sequence-to-sequence (seq2seq) model with a copy mechanism to predict a target sequence, which is concatenated of present and absent keyphrases. Therefore, the generative approach can predict both kinds of keyphrases. But these methods treat present and absent keyphrases equally, while these two kinds of keyphrase actually have different semantic properties.
As illustrated in Figure \ref{fig1:case-study}, all the present keyphrases are specific techniques, while the absent keyphrases are tasks or research areas.

Thus several integrated methods \cite{chen2019integrated, ahmad-etal-2021-select} try to perform multi-task learning on present keyphrase extraction (PKE) and absent keyphrase generation (AKG). 
By treating present and absent keyphrase prediction as different tasks, integrated methods clearly distinguish the semantic properties for these two kinds of keyphrases. 
But integrated models suffer from two limitations. Firstly, these approaches are not trained in an end-to-end fashion, which causes error accumulation in the pipeline. 
Secondly, integrated methods just adopt a bottom shared encoder to implicitly capture the latent semantic relation between PKE and AKG, while this relation is essential for the KP task.
As illustrated in Figure \ref{fig1:case-study}, the ground truth of PKE are specific techniques, which are all used for the ``singularity detection'' task in the ``traffic data analysis'' area.
Such semantic relation between PKE and AKG can bring benefits for KP.
Actually, semantic relations like ``technique-task-area'' between two tasks are common in the KP task.
However, these integrated methods are weak at modeling it.

To address these issues, we propose a novel end-to-end joint model, UniKeyphrase, which adopts a unified pretrained language model as the backbone and is fine-tuned with both PKE and AKG tasks. What's more, UniKeyphrase explicitly captures the mutual relation between these two tasks, which brings benefits for keyphrase prediction: present keyphrases can provide an overall sense about salient parts of the document for AKG, and absent keyphrases viewed as high-level latent topics of the document can also supply PKE with global semantic information.

Specifically, UniKeyphrase employs two mechanisms to capture the relation from model structure and training process, respectively. Firstly, stacked relation layer is applied to repeatedly fuse PKE and AKG task representations to explicitly model the relation between the two sub-tasks. In detail, we adopt a co-attention based relation network to model the co-influence. Secondly, a bag-of-words constraint is designed for UniKeyphrase, which aims to provide some auxiliary global information of the whole keyphrases set during training.

Experiments conducted on the widely used public datasets show that our method significantly outperforms mainstream generative and integrative models.\footnote{Code available on https://github.com/thinkwee/UniKeyphrase} The contributions of this paper can be summarized as follows:
\begin{itemize}
    \item We introduce a novel end-to-end framework UniKeyphrase for unified PKE and AKG.
    \item We design stacked relation layer (SRL) to explicitly capture the relation between PKE and AKG.
    \item We propose bag-of-words constraint (BWC) to explicitly feed global information about present and absent keyphrases to the model.
\end{itemize}

\section{Related Works}

\subsection{Keyphrase Extraction}
Most existing extraction approaches can be categorized into two-step extraction methods and sequence labeling approaches. Two-step extraction methods first identify a set of candidate phrases from the document by heuristics, such as essential n-grams or noun phrase~\cite{hulth-2003-improved}. Then, the candidate keyphrases are sorted and ranked to get predicted results. The scores can be learned by either supervised algorithms~\cite{nguyen2007keyphrase, medelyan2009human, lopez2010humb} or unsupervised graph ranking methods~\cite{mihalcea2004textrank, wan2008single}. For sequence labeling approaches, documents are fed to an encoder then the model learns to predict the likelihood of each word being a keyphrase~\cite{zhang2016keyphrase, alzaidy2019bi, sun2020joint}. 

\subsection{Keyphrase Generation}
Keyphrase generation focuses on predicting both present and absent keyphrases. \citet{meng2017deep} first propose CopyRNN which is a seq2seq
framework with attention and copy mechanism. Then a semi-supervised method for the exploitation of the unlabeled data is investigated by \citet{ye2018semi}. \citet{chen2018keyphrase} employ a review mechanism to reduce duplicates. \citet{chen2019guided} focus on leveraging the title information to improve keyphrases generation. The latent topics of the document are exploited to enrich features by \citet{wang2019topic}. \citet{zhao2019incorporating} utilize linguistic constraints to prevent model from generating overlapped phrases. \citet{chan2019neural} introduce a reinforcement learning approach for keyphrase generation.
\citet{chen2020exclusive} propose an exclusive hierarchical decoding framework to explicitly model the hierarchical compositionality of a
keyphrase set. \citet{yuan2020one} introduce a new model to generate multiple keyphrases as delimiter-separated sequences.

\subsection{Integrated Methods}
To explicitly distinguish the present and absent keyphrases, integrated extraction and generation approach have been applied to the KP task. \citet{chen2019integrated} aim at improving the performance of the generative model by using an extractive model.
\citet{ahmad-etal-2021-select} propose SEG-Net, a neural keyphrase generation model that is composed of a selector for selecting the salient sentences in a document, and an extractor-generator that extracts and generates keyphrases from the selected sentences. In contrast to these methods, our joint approach can explicitly capture the relation between extraction and generation in an end-to-end framework.

\section{Approach}
\begin{figure*}[!t]
\centering\includegraphics[width=1\textwidth]{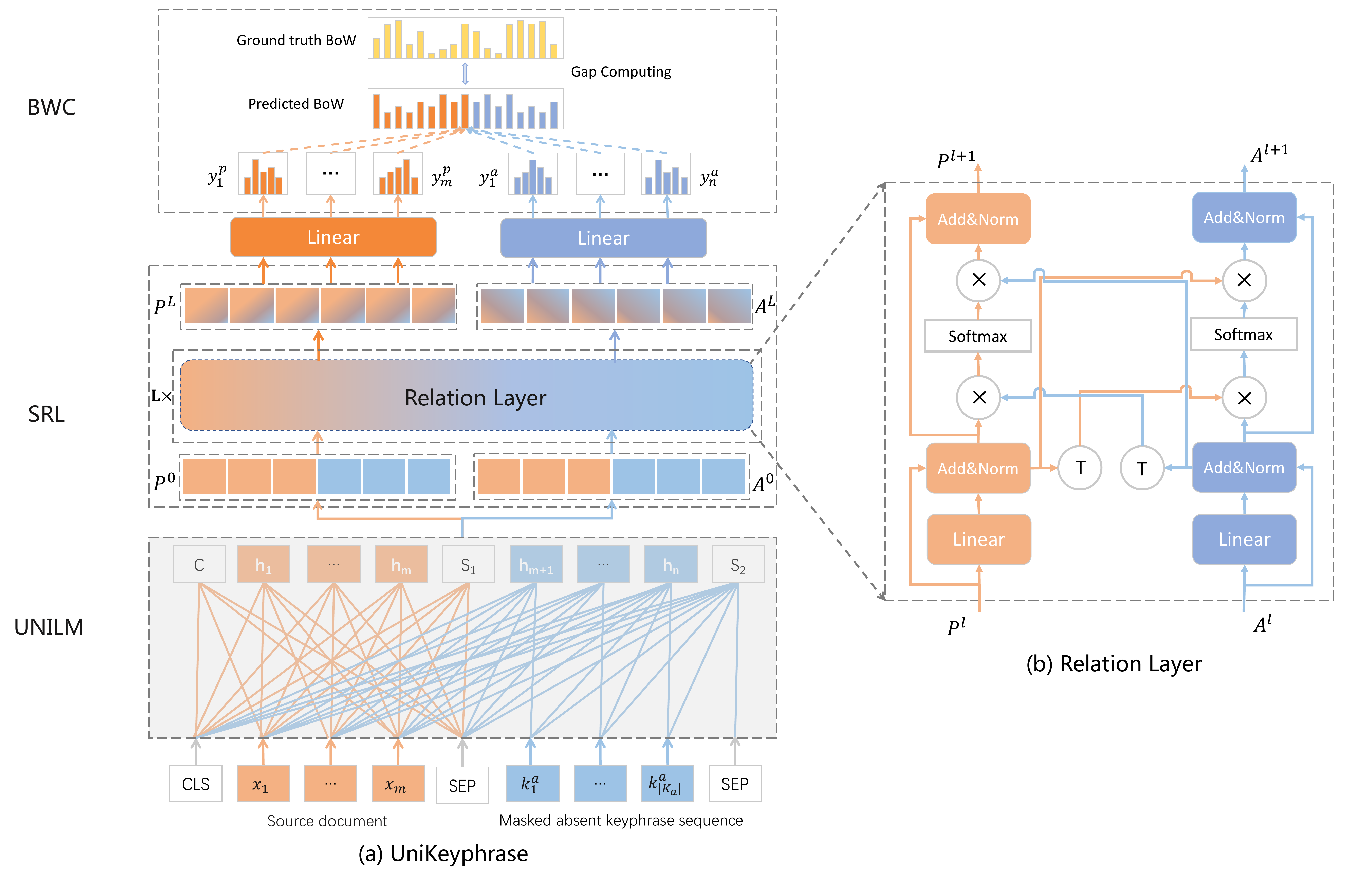}
\caption{The architecture of our model}
\label{fig1:model}
\end{figure*}

In this section, we describe the architecture of  UniKeyphrase. Figure \ref{fig1:model} gives an overview of UniKeyphrase, which consists of three components: extractor-generator backbone based on U{\small NI}LM, a stacked relation layer for capturing the relation between PKE and AKG, and bag-of-words constraint for considering the global view of two tasks in training. In the following sections, the details of UniKeyphrase are given.

\subsection{Extractor-Generator Backbone}
Given a document ${\textbf{X}} = \{x_{1}, ..., x_{m}\}$, KP aims at obtaining a keyphrase set 
${\textbf{K}} = \{k_{1}, ..., k_{|K|}\}$. Naturally, ${\textbf{K}}$ can be divided into present keyphrase set ${\textbf{K}_{p}} = \{k_{1}^{p}, ..., k_{|{K}_{p}|}^{p}\}$ and absent keyphrase set ${\textbf{K}_{a}} = \{k_{1}^{a},, ..., k_{|{K}_{a}|}^{a}\}$ by judging whether keyphrases appear exactly in the source document. UniKeyphrase decomposes the KP into PKE and AKG, and jointly learns two tasks in an end-to-end framework.

UniKeyphrase treats PKE as a sequence labeling task and AKG as a text generation task. 
To jointly learn in an end-to-end framework, UniKeyphrase adopts U{\small NI}LM \cite{dong2019unified} as the backbone network. 
U{\small NI}LM is a pre-trained language model, which can perform sequence-to-sequence prediction by employing a shared transformer network and utilizing specific self-attention masks to control what context the prediction conditions on. 

As shown in Figure \ref{fig1:model}, with a pre-trained U{\small NI}LM layer, the contextualized representation for the source document can attend to each other from both directions, which is convenient for PKE.
While the representation of the target token can only attend to the left context, as well as all the tokens in the source document, which can be easily adapted to AKG.

Specifically, for a document ${\textbf{X}}$, 
all absent keyphrases will be concatenated as a sequence. Then we randomly choose tokens in this sequence, and replace them with the special token [MASK]. The masked sequence is defined as $\textbf{K}_{a}^{m}$. We further concatenate document $\textbf{X}$ and $\textbf{K}_{a}^{m}$ with [CLS] and [SEP] tokens as the input sequence:
\def\CLS{\mathop{\rm [CLS]}}
\def\SEP{\mathop{\rm [SEP]}}
\begin{gather}
\textbf{I} = \{ \CLS \textbf{X} \SEP \textbf{K}_{a}^{m} \SEP \} 
\end{gather}

Afterwards, we feed input sequence into U{\small NI}LM and obtain output hidden state $\textbf{H}$:
\def\UNILM{\mathop{\rm U{\small NI}LM}}
\begin{gather}
\textbf{H} = \UNILM(\textbf{I})
\end{gather}
the hidden state ${\textbf{H}}$ = \{$h_{1}$, ...,$h_{T}$\} ($T$ is the number of input tokens in the U{\small NI}LM) will be used as the input of stacked relation layer for jointly modeling PKE and AKG. 

\subsection{Stacked Relation Layer}
Based on the U{\small NI}LM, we can obtain the output hidden ${\textbf{H}}$.
Instead of directly using the U{\small NI}LM hidden for PKE and AKG, we use the SRL to explicitly model the relation between these two tasks. Actually, modeling the cross-impact and interaction between different tasks in joint model is a common problem \cite{qin2020dcr, qinetal2020agif, qinetal2019stack}.

Specifically, SRL takes the initial shared representations ${\textbf{P}}^{0}$ = ${\textbf{A}}^{0}$ = \{$h_{1}$, ...,$h_{T}$\} as input and aims to obtain the finally task representations ${\textbf{P}^{L}}$ and ${\textbf{A}^{L}}$ ($L$ is the number of stacked layers), which consider the cross-impact between PKE and AKG. 
Besides, SRL can be stacked to repeatedly fuse PKE and AKG task representations for better capturing mutual relation.

Formally, given the $l^\text{th}$ layer inputs ${\textbf{P}}^{l}$ = \{$p_{1}^{l}$, ...,$p_{T}^{l}$\} and  ${\textbf{A}}^{l}$ = \{$a_{1}^{l}$, ...,$a_{T}^{l}$\},
stacked relation layer first apply two linear transformations with a ReLU activation over the input to make them more task-specific, which can be written as follow:

\def\LN{\mathop{\rm LN}}
\def\Max{\mathop{\rm max}}
\begin{align}
{ \textbf{P}}^{l^{'}}  &=\LN ({ \textbf{P}}^{l} + \Max(0, \textbf{W}_{P}^{l} \textbf{P}^{l} + \textbf{b}_{P}^{l})) \\
{ \textbf{A}}^{l^{'}}  &=\LN ({ \textbf{A}}^{l} + \Max(0, \textbf{W}_{A}^{l} \textbf{A}^{l} + \textbf{b}_{A}^{l}))    
\end{align}
where $ \rm{LN} $ represent the layer normalization function \cite{ba2016layer}.

Then the relation between the two tasks will be integrated base on task-specific representations.
In this paper, we adopt co-attention relation networks.
Co-Attention is an effective approach to model the important information of correlated tasks.
We extend the basic co-attention mechanism from token level to task representations level.
It can produce the PKE and AKG task representations considering each other. Therefore, we can transfer useful mutual information between two tasks.
The process can be formulated as follows:

\def\Softmax{\mathop{\rm softmax}}
\begin{gather}
{ \textbf{P}}^{l+1}= { \LN (\textbf{P}}^{l^{'}} + \Softmax ({ \textbf{P}}^{l^{'}}({ \textbf{A}}^{l^{'}}) ^\top) { \textbf{A}}^{l^{'}}) \\
{ \textbf{A}}^{l+1} = { \LN (\textbf{A}}^{l^{'}} + \Softmax ({ \textbf{A}}^{l^{'}} ({ \textbf{P}}^{l^{'}}) ^\top) { \textbf{P}}^{l^{'}})
\end{gather}
where ${\textbf{P}}^{l+1}$ = \{$p_{1}^{l+1}$, ...,$p_{T}^{l+1}$\} and  ${\textbf{A}}^{l+1}$ = \{$a_{1}^{l+1}$, ...,$a_{T}^{l+1}$\} are the $l^\text{th}$ layer updated representations.

After stacked relation layer, we can obtain the outputs ${\textbf{P}}^{L}$ = \{$p_{1}^{L}$, ...,$p_{m}^{L}$\} and $\textbf{A}^{L}$ = \{$a_{1}^{L}$, ...,$a_{n}^{L}$\}. We then adopt separate decoders to perform PKE and AKG by using the task representations of corresponding position , which can be denoted as follows: 
\def\softmax{\mathop{\rm softmax}}
\begin{gather}
	\textbf{y}_{i}^{p} = \softmax (\textbf{W}^{p}{ p }_{i}^{L} + \textbf{b}^{p}) \\
\textbf{y}_{j}^{a} = \softmax (\textbf{W}^{a}{a }_{j}^{L} + \textbf{b}^{a})
\end{gather}
where $\textbf{y}_{i}^{p} $ and $\textbf{y}_{j}^{a}$ are the predicted distribution for present keyphrase and absent keyphrase respectively; $\textbf{W}^{p}$ and $\textbf{W}^{a}$ are transformation matrices; $\textbf{b}_{p}$ and $\textbf{b}_{a}$ are bias vectors.

\subsection{Bag-of-Words Constraint}
UniKeyphrase divides the KP task into two sub-tasks, PKE and AKG. These two sub-tasks are optimized separately, which lacks the awareness of global information about the total keyphrase set. Such global information can be the amount of all keyphrases or the common words between present and absent keyphrases. Bag of words (BoW) is a suitable medium for describing this information. In this paper, we feed global information to UniKeyphrase by constructing constraints based on the BoW of keyphrases. The word count in BoW can provide guidance about task relation for PKE and AKG training in a global view.

Specifically, we calculate the gap between the model predicted keyphrase BoW and ground truth keyphrase BoW, then add it into the loss. Hence UniKeyphrase can get a global view of keyphrases allocation and adjust two tasks during training.

We first collect present and absent keyphrase BoW from model. For present keyphrases, since PKE is a sequence labeling task, we collect all words that labeled as keyphrases, and construct present predicted BoW $V^{p}$. We use the sum of corresponding label probabilities as the count of word $w$ in $V^{p}$:
\begin{gather}
V^{p}(w) = \sum_{i \in \mathcal{I} _w} max(\textbf{y}_i^p) 
\end{gather}
where $y_i^p$ denotes all predicted label probabilities at time step $i$. $\mathcal{I} _w$ is all position of word $w$ in document. Maximum operation is used for selecting the probability of predicted label. For absent keyphrase, the generation probability of all steps are accumulated as predicted absent BoW $V^{a}(w)$.
\begin{gather}
V^{a}(w) = \sum_{j=1}^N \textbf{y}_j^a(w)  
\end{gather}
After acquiring the predicted present and absent keyphrase BoW, we concatenate these two parts as the total predicted BoW $V$, then calculate the error compared with ground truth BoW $\hat{V}$. To reserve the word count information, we use Mean Square Error (MSE) function:
\begin{gather}
\mathcal{L}_{BoW} = \frac{1}{|\mathcal{V}|}\sum_{w \in \mathcal{V}} (V(w) - \hat{V}(w))^2 
\end{gather}
It is worth noting that $\mathcal{V}$ is the collection of words that make up the ground truth keyphrases and predicted keyphrases. 
So the BWC only affects a small subset of the whole vocabulary for each sample. This can help reduce the noise and stabilize the training process. 

In practice we increase the weight of BWC logarithmically from zero to a defined maximum value $w_m$, the weight of BWC on $t$ step can be denoted as follows:
\begin{gather}
w_{BoW}(t) = log(\frac{e^{w_m} - 1}{t_{total}} t + 1)
\end{gather}
where $t_{total}$ is the total step of training. The reason to adjust the weight is the same as~\citet{ma2018bag}. The BWC should take effect when predicted results are good enough. Therefore we first assign a small weight to BWC at the initial time, and gradually
increase it when training.

\subsection{Training}
For the PKE task, objection is formulated as:
\begin{gather}
\mathcal {L}_{PKE} = - \sum_{i= 1}^{M}  \sum_{c= 1}^{C} w_{c} \hat{{\bf{y}}}_{i}^{(c, p)} \log \left( {\bf{y}}_{i} ^ {(c, p)} \right) 
\end{gather}
where $M$ refers to the length of document, $C$ refers to the number of label, $w_{c}$ is the loss weight for the positive label.
$\hat{{\bf{y}}}_{i}^{p}$ refers the gold label.

For the AKG task, training objection is to
maximize the likelihood of masked tokens, which is formulated as: 

\begin{gather}
\mathcal{L}_{AKG} = -\sum_{i=1}^{N} \sum_{j = 1}^{V_s} \hat{{\bf{y}}}_{i}^{(j, a)} \log \left( {\bf{y}}_{i} ^ {(j, a)}\right)
\end{gather}
where $N$ refers to the number of masked tokens, $V_s$ refers to the size of vocabulary.
$\hat{{\bf{y}}}_{i}^{a}$ refers the ground-truth word.

Considering the BWC, the overall loss of UniKeyphrase is formulated as:
\begin{gather}
\mathcal{L} = \mathcal{L}_{PKE} + \mathcal{L}_{AKG} + w_{BoW}\mathcal{L}_{BoW}   
\end{gather}

\section{Experiments}

\subsection{Datasets and Evaluation}
We follow the widely used setup of the deep KP task: train, validation and test on the \texttt{KP20K}~\cite{meng2017deep} dataset, and give evaluation on three more benchmark datasets: \texttt{NUS}~\cite{nguyen2007keyphrase}, \texttt{INSPEC}~\cite{hulth-2003-improved} and \texttt{SEMEVAL}~\cite{kim2010semeval}. 
We follow the pre-process, post-process, and evaluation setting of  ~\citet{meng2017deep,meng2019ordermatters, yuan2020one}\footnote{we follow the official GitHub repository to prepare datasets and evaluation scripts which are available on https://github.com/memray/OpenNMT-kpg-release.}. Specifically, we use the partition of present and absent provided by ~\citet{meng2017deep} and calculate $F_1$@5 and $F_1$@M (use all predicted keyphrases for $F_1$ calculation) after stemming and removing duplicates.

\subsection{Experimental Setup}
\textbf{Setting}: 
We reuse most hyper-parameters from pre-trained U{\small NI}LM\footnote{we use the official provided pre-trained model, which is available on https://unilm.blob.core.windows.net/ckpt/unilm1-base-cased.bin.}.
The layer number of SRL is set to 2.
We use $w_m = 1.0$ when adjusting the weight of BWC.
PKE loss weights $w_c$ for the positive label is set to 5.0.
we set batch size to 256, and maximum length to 384.
During decoding, we use beam search for AKG, and beam size is set as 5.
We train our model on the training set for 100 epochs.
It takes about 40 minutes per epoch to train UniKeyphrase on 8 Nvidia Tesla V100 GPU cards with mixed-precision training. 
More details are provided in Appendix B.

\begin{table*}[htbp]
\centering
\scalebox{0.75}{
\begin{tabular}{c|cccccccccc}
\toprule[1pt]
\multirow{2}{*}{Type} &
  \multirow{2}{*}{Model} &
  \multicolumn{2}{c}{KP20k} &
  \multicolumn{2}{c}{NUS} &
  \multicolumn{2}{c}{SemEval} &
  \multicolumn{2}{c}{Inspec} \\
                             &                   & $F_1$@5 & $F_1$@M & $F_1$@5 & $F_1$@M & $F_1$@5 & $F_1$@M & $F_1$@5 & $F_1$@M  \\ \hline
\multirow{6}{*}{Generative} & CatSeq            & 29.1 & 36.7 & 32.3 & 39.7 & 24.2 & 28.3 & 22.5 & 26.2 \\
                             & CatSeqTG          & 29.2 & 36.6 & 32.5 & 39.3 & 24.6 & 29.0 & 22.9 & 27.0 \\
                             & CatSeq(TRM)       & 29.1 & 36.8 & 32.8 & 40.5 & 24.5 & 28.8 & 22.5 & 26.4 \\
                             & CatSeqD           & 28.5 & 36.3 & 32.1 & 39.4 & 23.3 & 27.4 & 21.9 & 26.3 \\
                             & ExHiRD-h          & 31.1 & 37.4 & -- & -- & 28.4 & \textbf{33.5} & 25.3 & \textbf{29.1} \\ 
                            \hline
\multirow{2}{*}{Integrated}   & KG-KE-KR-M      & 31.7 &  --  & 28.9 &  -- & 20.2 &  --  & 25.7 & --   \\
                             & SEG-NET           & 31.1 & \textbf{37.9} & 39.6 & \textbf{46.1} & 28.3 & 33.2 & 21.6 & 26.5 \\ \hline
           Joint            & UniKeyphrase      &  \textbf{34.7}  & 35.2 &  \textbf{41.5}  & 44.3 & \textbf{30.2} & 32.2 & \textbf{26.0}  & 28.8 \\ 
\bottomrule[1pt]
\end{tabular}
}
\caption{Results on present keyphrase prediction.}
\label{tab:result-present}
\end{table*}

\textbf{Baselines}: We compare two kinds of strong baselines (generative, integrated) to give a comprehensive evaluation on the performance of UniKeyphrase.
\begin{itemize}
    \item \textbf{Generative}: Generative models can predict both present and absent keyphrases under the seq2seq framework. CatSeq~\cite{yuan2020one} is the classic setting of keyphrase seq2seq model. We report the performance of CatSeq and various improved models on it, including CatSeqTG~\cite{chen2019guided}, CatSeq (TRM)~\cite{ahmad-etal-2021-select} and CatSeqD~\cite{yuan2020one}. A recently released model is also included for comparing, which is ExHiRD-h~\cite{chen2020exclusive}.
    \item \textbf{Integrated}: Integrated model often combine multiple modules to perform extractive and abstractive tasks. But they are not end-to-end. Two latest integrated models are recorded for comparison. including KG-KE-KR-M~\cite{chen2019integrated} and SEG-NET~\cite{ahmad-etal-2021-select}
\end{itemize}

\subsection{Main Results}
In this section, we show the experimental results of the baseline methods and our model on present keyphrase extraction and absent keyphrase generation. Besides, we also study the average number of unique predicted keyphrases per document to further show the advantages of our model.

\subsubsection{Present and Absent Keyphrase Prediction}
The present and absent keyphrase prediction performance of all methods are shown in Table \ref{tab:result-present} and Table \ref{tab:result-absent}. From the results, we can find that our joint framework outperforms most state-of-the-art generative baseline by a significant margin, especially on absent keyphrase generation, which demonstrates the effectiveness of our UniKeyphrase.
We notice that the UniKeyphrase does not perform well on $F_1$@M for present keyphrase extraction. One potential reason is that UniKeyphrase predicts more than other baselines, which makes it has the potential to predict more reasonable but not-ground-truth keyphrases.
\begin{table*}[]
\centering
\scalebox{0.75}{
\begin{tabular}{c|ccccccccc}
\toprule[1pt]
\multirow{2}{*}{Type} &
  \multirow{2}{*}{Model} &
  \multicolumn{2}{c}{KP20k} &
  \multicolumn{2}{c}{NUS} &
  \multicolumn{2}{c}{SemEval} &
  \multicolumn{2}{c}{Inspec} \\
                             &                   & $F_1$@5 & $F_1$@M & $F_1$@5 & $F_1$@M & $F_1$@5 & $F_1$@M & $F_1$@5 & $F_1$@M \\ \hline
\multirow{6}{*}{Generative} & CatSeq            & 1.5  & 3.2  & 1.6  & 2.8  & 2.0  & 2.8  & 0.4  & 0.8  \\
                             & CatSeqTG          & 1.5  & 3.2  & 1.1  & 1.8  & 1.9  & 2.7  & 0.5  & 1.1  \\
                             & CatSeq(TRM)       & 1.5  & 3.1  & 1.1  & 1.8  & 1.9  & 2.7  & 0.5  & 0.9 \\
                             & CatSeqD           & 1.5  & 3.1  & 1.5  & 2.4  & 1.6  & 2.4  & 0.6  & 1.1 \\
                             & ExHiRD-h          & 1.6  & 3.2  & --  & --  & 1.7  & 2.5  & 1.1  & 2.2 \\ 
                            \hline                
\multirow{2}{*}{Integrated}    & KG-KE-KR-M\tablefootnote{Reports from~\citet{yuan2020one}, which do not report absent metrics for this model. The original paper also does not give detailed numbers.}  &  --  &  --  &  --  &  --   &  --  &  --  &  --  &  --  \\
                             & SEG-NET           & 1.8  & 3.6  & 2.1 & 3.6 & 2.1 & 3.0  & 0.9 & 1.5 \\ \hline
\multirow{2}{*}{Joint}            & UniKeyphrase      & 3.2  & 5.8   & 2.6  &  3.7  & 2.2 & 2.9 & 1.2 & 2.2  \\ 
                                       & UniKeyphrase(beam=4)      & \textbf{4.6}  & \textbf{6.8}   & \textbf{4.5}  &  \textbf{5.6}  &  \textbf{4.5}  &  \textbf{5.2}  & \textbf{2.6}  & \textbf{3.6}   \\ 
\bottomrule[1pt]
\end{tabular}
}
\caption{Results on absent keyphrase prediction.}
\label{tab:result-absent}
\end{table*}

\begin{table}[!t]
\centering
\resizebox{0.98\columnwidth}{!}{
\begin{tabular}{l| c c| c c| c c}
\toprule[1pt]
  \multirow{2}{*}{\textbf{Model}} & \multicolumn{2}{c|}{\textbf{Inspec}} &
  \multicolumn{2}{c|}{\textbf{SemEval}} &
  \multicolumn{2}{c}{\textbf{KP20k}}
  \\
  & \text{\#}PK & \text{\#}AK & \text{\#}PK & \text{\#}AK & \text{\#}PK & \text{\#}AK \\
  \hline
 Ground Truth        & 7.64 & 2.10 & 6.28 & 8.12 & 3.32 & 1.93 \\
 \hline
 Transformer   & 3.17 & 0.70 & 3.24 & 0.67 & 3.44 & 0.58 \\
 catSeq        & 3.33 & 0.58 & 3.45 & 0.64 & 3.70 & 0.51 \\
 catSeqD       & 3.33 & 0.58 & 3.47 & 0.63 & 3.74 & 0.50 \\
 catSeqCorr    & 3.07 & 0.53 & 3.15 & 0.62 & \textbf{3.36} & 0.50 \\
 ExHiRD-h      & 4.00 & 1.50 & 3.65 & 0.99 & 3.97 & 0.81\\
 SEG-NET       & - & - & - & - & 3.79 & 1.14 \\
 \hline
 UniKeyphrase  & \textbf{5.61} & \textbf{1.77} & \textbf{5.60} & \textbf{1.52}  & 6.07 & \textbf{1.75}  \\
\toprule[1pt]
\end{tabular}
}
\caption{Results of average numbers of predicted unique keyphrases. ``\text{\#}PK'' and ``\text{\#}AK'' are the number of present and absent keyphrases respectively. \textbf{Bold} denotes the prediction closest to the ground truth.
}
\label{table:number of predicted keyphrases}
\end{table}

\subsubsection{Number of Predicted Keyphrases}
The number of predicted keyphrases indicates the model’s understanding of input documents. 
From the previous work \cite{chen2020exclusive}, we find the average number of unique predicted keyphrases per document is much lower than the gold average keyphrase number in most datasets. 
The number of unique keyphrases predicted by UniKeyphrase and baselines is compared in Table \ref{table:number of predicted keyphrases}. We can find that UniKeyphrase predicts more (especially in absent keyphrases) than baseline methods, which is closer to ground truth.
Meanwhile, we find UniKeyphrase leads to predict more keyphrases than the ground-truth (especially on KP20k). We leave solving the over prediction keyphrases problem as our future work.

\begin{table}[!t]
\centering
\scalebox{0.85}{
\begin{tabular}{l|ccc}
\toprule[1pt]
\multirow{2}{*}{Model} &
  \multicolumn{3}{c}{Dataset} \\
                             & KP20k & Inspec & NUS  \\ \hline
UniKeyphrase            &   35.2 &   28.8   &  44.3   \\ \hline
w/o SRL    &    35.1   &   26.2    &   41.9        \\ 
w/o BWC      &  34.3   &  29.1  &  43.2 \\ 
\bottomrule[1pt]
\end{tabular}
}
\caption{Ablation study of present F1@m on Three dataset}
\label{tab:ablation-study}
\end{table}

\subsection{Ablation Study}
In this section, we check the improvement brought by SRL and BWC. 
Several ablation experiments are conducted to analyze the effect of different components. 
The ablation experiment on three datasets are shown in Table \ref{tab:ablation-study}.
The results show the effectiveness of different components of our method to the final performance.

\textbf{Effectiveness of stacked relation layer}: In this setting, we conduct experiments on the multi-task framework where PKE and AKG promote each other only by the hidden state of U{\small NI}LM, From the result, we can see that the performance drops both in present keyphrase and absent keyphrase without stacked relation layer. This demonstrates that explicitly modeling the relation between PKE and AKG with stacked relation layer can benefit them effectively.

\textbf{Effectiveness of bag-of-words constraint}: In this setting, we remove our bag-of-words constraint and there is no global constraint for two tasks. The results show a drop in KP performance, indicating that capturing the global constraint of the result by BWC is effective and important for our method.

\subsection{Analysis}
\subsubsection{SRL Analysis}
To better understand the SRL module, we analyze the impact of stacked layers and give a visualization of the inner state of SRL.

\textbf{Analysis of SRL Layer Number}: We explore the impact of the stack number of relation network.
The comparison of total keyphrase prediction
result, which regardless of the present or absent of keyphrases, are shown in Table \ref{tab:task_comparison}. We can find that setting deeper layers could generally result in better performance when the number of stacked layers is less than three, which proves the effectiveness of stacked layers. 
It is worth noting that when the number of stacked layers is larger than two, the KP performance drops. 
We suppose that when the relation network becomes deeper, the over-interaction will lose the diversity of two task representations.

\begin{table}[t]
\small
\centering
\scalebox{0.85}{
\begin{tabular}{c|ccc}
\toprule[1pt]
\multirow{2}{*}{Model}           & \multicolumn{3}{c}{Dataset}                     \\ \cline{2-4} 
                                 & Inspec         & NUS            & SemEval        \\ \hline
UNILM based keyphrase generation & 22.02          & 23.95          & 17.91          \\
UniKeyphrase with 0 layer SRL    & 20.60          & 27.27          & 20.36          \\
UniKeyphrase with 1 layer SRL    & 20.69          & 26.41          & 19.41          \\
UniKeyphrase with 2 layer SRL    & \textbf{26.56} & \textbf{28.56} & \textbf{20.80} \\
UniKeyphrase with 3 layer SRL    & 23.40          & 28.25          & 20.60          \\ 
\bottomrule[1pt]
\end{tabular}
}
\caption{Total keyphrase prediction on SemEval dataset by different setting. UniKeyphrase with 0 layer SRL means UniKeyphrase without SRL module.}
\label{tab:task_comparison}
\end{table}

\begin{figure}[ht]
\centering\includegraphics[width=0.4\textwidth]{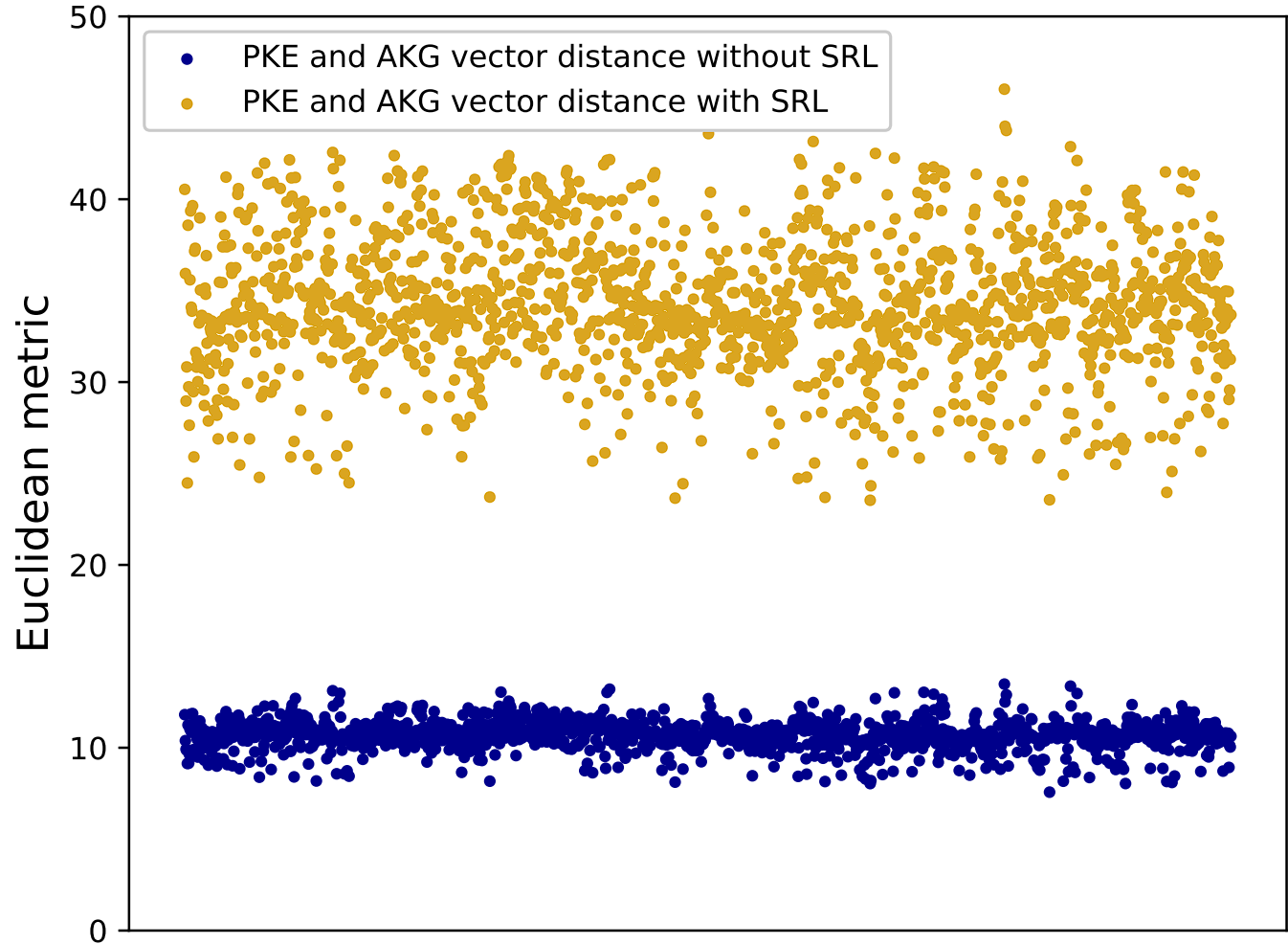}
\caption{Distance between PKE representation and AKG representation on different settings.}
\label{fig:visualization_analysis}
\end{figure}

\textbf{Visualization Analysis for SRL}: To better understand what the SRL network has learned, we compare the distance between the PKE representation and AKG representation in different settings.
In detail, we randomly sample 2000 pairs of PKE representation vector and AKG representation vector on different positions from test data and compute euclidean metric in each pair.
As shown in Figure \ref{fig:visualization_analysis}, the blue points mean the Euclidean metric between PKE and AKG representation vector without SRL layer, while the yellow points mean the Euclidean metric with SRL layer.

From the Figure \ref{fig:visualization_analysis}, we can find that the blue points are under the yellow points, which means the PKE and AKG representation vector without SRL is more similar. In other words, SRL has learned the task-specific representation.
Also, the blue points are denser than the yellow points, which means the PKE and AKG representation with SRL is more diverse than the one without SRL on different samples.

\subsubsection{BWC Analysis}
\begin{figure}[]
\centering\includegraphics[width=0.4\textwidth]{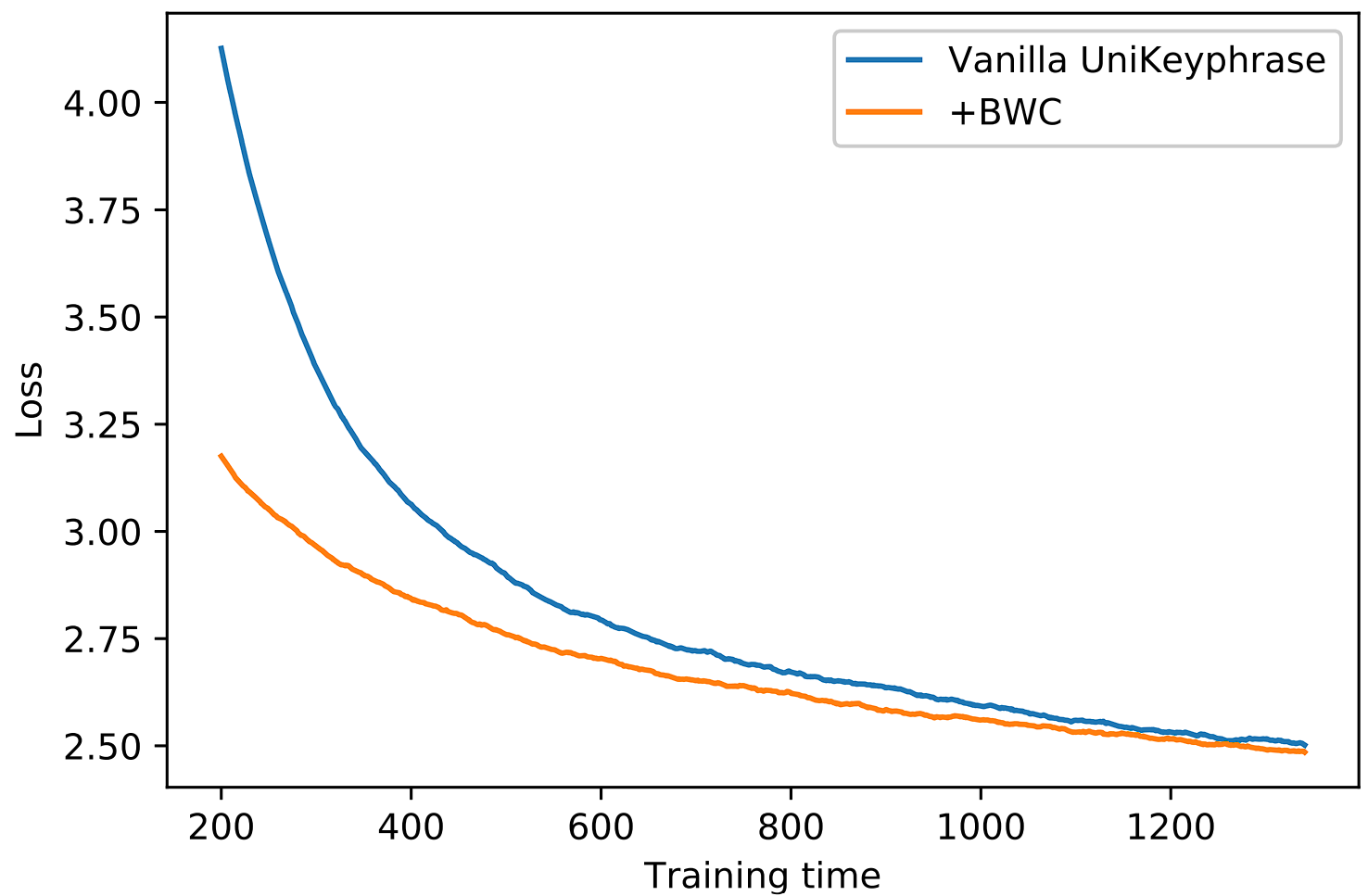}
\caption{BWC's influence on total training loss (sequence labeling + text generation).}
\label{fig:bwc_loss}
\end{figure}
\textbf{Loss Compare}: From Figure \ref{fig:bwc_loss} we can see that the original total loss (labeling and generation) drops more with the help of BWC compared to the vanilla model. BWC actually is an enhancement on the original supervised signal from a global view. It guides the model to learn how many to predict and how to allocate present and absent keyphrases, while original loss only teaches what to predict in each position.

\begin{figure}[]
\centering\includegraphics[width=0.4\textwidth]{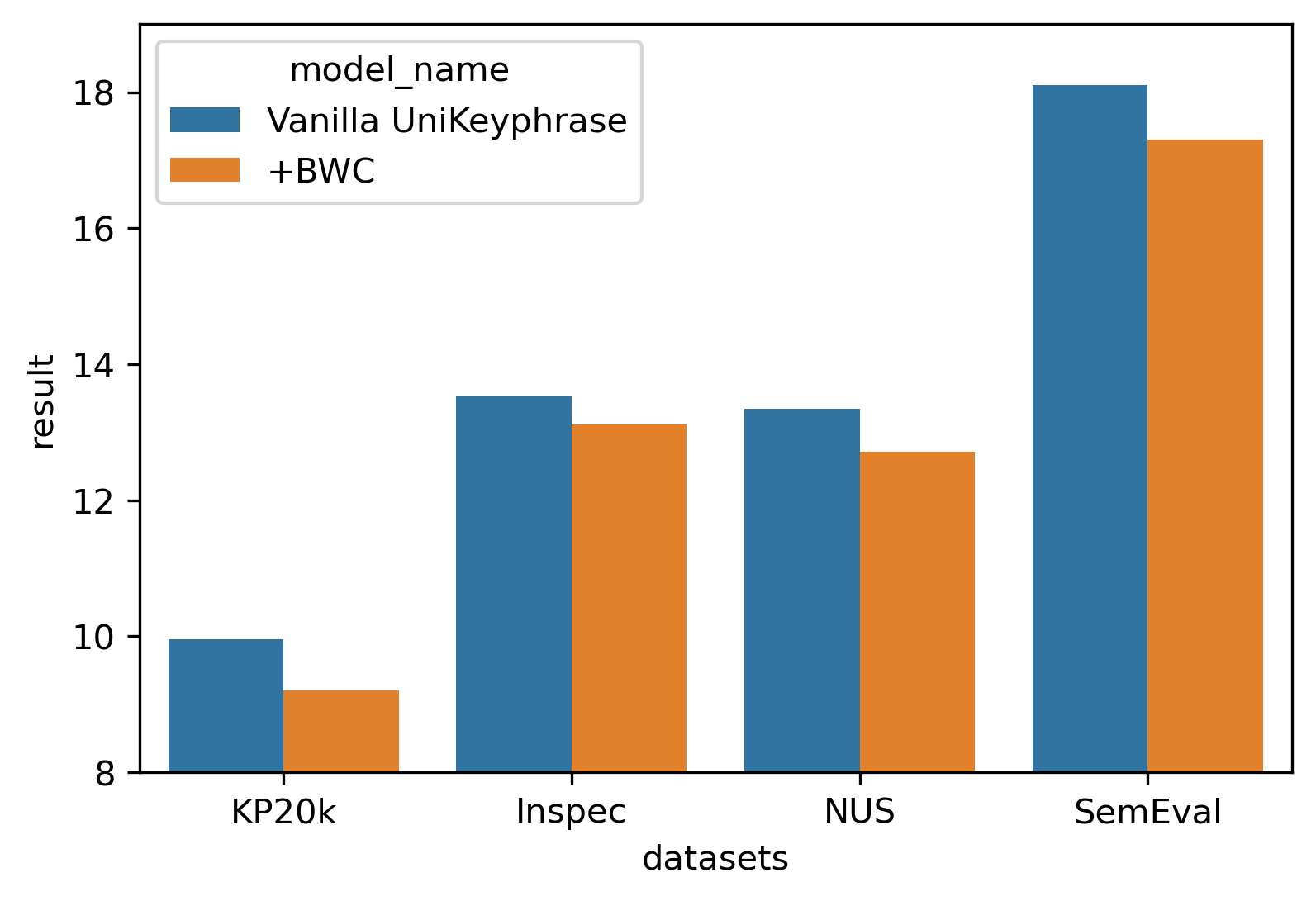}
\caption{Bag-of-words Error comparison between vanilla and BWC.}
\label{fig:bow_diff}
\end{figure}
\textbf{Bag-of-words Error}: We also calculate the bag-of-words Error between ground truth and model predicted keyphrases, which is how many tokens are incorrectly predicted. As shown in Figure \ref{fig:bow_diff}, UniKeyphrase with BWC achieves lower BoW Error compared with the vanilla model. It proves that BWC successfully guides the model to learn a better BoW allocation.

\subsubsection{Joint Framework Analysis}
In our UniKeyphrase model, we adopt pre-trained model U{\small NI}LM for KP. 
So it is necessary to check that the gain on metrics of our proposed joint framework is not just come from the pre-trained model. In this section, we compare UniKeyphrase with directly using the pre-trained U{\small NI}LM to perform generative KP.

Specifically, we train a sequence to sequence model for KP based on U{\small NI}LM. Results are shown in Table \ref{tab:task_comparison}. From the results, we find that all of the joint models with SRL can further outperform the generative method based on U{\small NI}LM, demonstrating that the improvement of KP mainly come from our joint framework instead of pre-trained U{\small NI}LM. We notice that the UniKeyphrase without SRL does not outperform the generative method based on U{\small NI}LM, which show the significance of modeling the relation between the two sub-tasks in our joint framework.

\section{Conclusion and Future Work}
This paper focuses on explicitly establishing an end-to-end unified model for PKE and AKG.
Specifically, we propose UniKeyphrase, which contains stacked relation layer to model the interaction and relation between the two sub-tasks.
In addition, we design a novel bag-of-words constraint for jointly training these two tasks.
Experiments on benchmarks show the effectiveness of the proposed model, and more extensive analysis further confirms the correlation between two tasks and reveals that modeling the relation explicitly can boost their performance.

Our UniKeyphrase can be formalized as a unified framework of NLU and NLG tasks. It is easy to transfer it to other extraction-generation NLP tasks. In the future, we will explore to adopt our framework to more scenarios.

\section*{Acknowledgments}
Lei Li were supported by Beijing Municipal Commission of Science and Technology [grant number Z181100001018035]; Engineering Research Center of Information Networks, Ministry of Education; BUPT Jinan Institute; Beijing BUPT Information Networks Industry Institute Company Limited.

\bibliographystyle{acl_natbib}
\bibliography{anthology,acl2021}

\clearpage
\newpage
\appendix
\section{Dataset Statistics}
\begin{table}[h]
\centering
\resizebox{1\columnwidth}{!}{
\begin{tabular}{l|l c c c c}
\toprule[1pt]
  \textbf{Type} & \textbf{Dataset} &  \textbf{\#Examples} & \textbf{Max/Avg \#Tokens} & \textbf{Max/Avg \#Sentences} \\
  \hline
  \multirow{4}{*}{Test} & Inspec & 500 & 387.0 / 138.4 & 27.0 / 6.7 & \\
  & NUS & 211 & 384.0 / 185.6 & 16.0 / 8.4 & \\
  & SemEval & 100 & 415.0 / 208.0 & 18 / 8.8 & \\
  & KP20k & 20000 & 1116.0 / 178.9 & 70.0 / 8.1 & \\
  \hline
  Validation & KP20k & 20000 & 1862.0 / 179.2 & 120 / 8.2 & \\
  \hline
  Train & KP20k & 514154 & 2924 / 177.9 & 284 / 8.2 & \\
\bottomrule[1pt]
\end{tabular}
}
\caption{Summary of the dataset used in experiments. ``\text{\#}Examples'' means the number of sample. ``\text{\#}Tokens'' means the number of token. ``\text{\#}Sentences'' means the number of sentence.
}
\label{table:dataset_statistic}
\end{table}

Relevant statistics about the dataset used in this paper is shown in Table \ref{table:dataset_statistic}.

\section{Experimental Details}
The BWC does not bring extra parameters, hence the trainable parameters of UniKeyphrase come from U{\small NI}LM and SRL. We use the base version of U{\small NI}LM, which contains about 110M parameters. Follow U{\small NI}LM, our model is implemented using PyTorch. The learning rate is 1e-5 and the proportion of warmup steps is 0.1. The masking probability of absent keyphrase sequence is 0.7.
For the SRL module, dropout is applied to the output of each layer for regularization, the dropout rate is 0.5. In this paper, we try to set the number of layer by 2,3,4 and choose the best based on validation. 
For all experiments in this paper, we choose the model that performs best on the KP20k validation dataset.

\section{Preprocess}
The input of UniKeyphrase is the same as BERT, which applies wordpiece tokenizer on raw sentences. So we use the ``BIXO" labeling method, where B and I stand for Beginning and Inside of a word in keyphrases, and O denotes any token that Outside of any keyphrase. For any sub-word token in keyphrases(which starts with `\#\#' in processed input) we use X to label it. For example, ``voip conferencing system'' will be tokenized into ``v \#\#oi \#\#p con \#\#fer \#\#encing system'' and be labeled as ``B X X I X X I''. We concatenate all the tokenized absent keyphrases into one sequence using a special delimiter “ ; ”. An example of absent keyphrase sequence will like ``peer to peer ; content delivery ; t \#\#f \#\#rc ; ran \#\#su \#\#b".

\section{Case Study}
\begin{figure}[!t]
\centering
\includegraphics[width=0.4\textwidth]{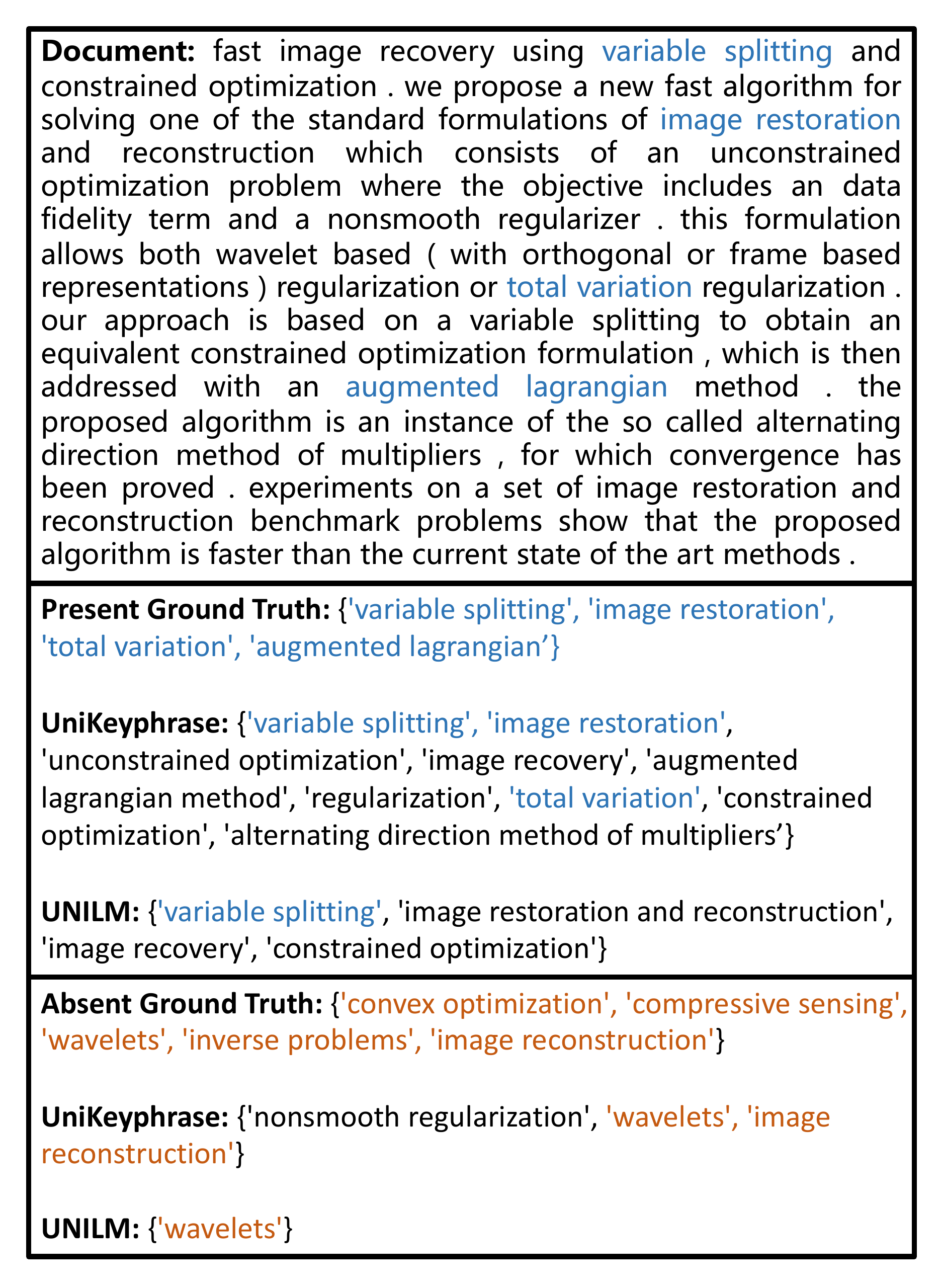}
\caption{Case study.}
\label{fig1:case-study-appendix}
\end{figure}
We give a case on the KP20k testset in Figure \ref{fig1:case-study-appendix}. We compare with the original U{\small NI}LM since our joint models are based on its implementation. Blue and red denote correct present and absent keyphrases, respectively. As shown in Figure \ref{fig1:case-study-appendix}, UniKeyphrase successfully catches the deep semantic relation similar to the case in the introduction and gives more accurate results(predicts some applications like "image restoration" or "image reconstruction"). 

\section{Evaluation Details}
We use $F_1$@5 and $F_1$@M as evaluation metric. Following previous works, we pad the result when number of predicted keyphrases is less than 5 when calculating $F_1$@5. For calculating $F_1$@5, since there is no explicit rank score for each predicted keyphrase, we calculate the rank score as follows:

\textbf{Present}: we calculate the average predicted label probabilities of all tokens in a keyphrase as the score. We tried several other scoring strategies as the score. The results show no significant difference(less than 0.1\%).

\textbf{Absent}: following previous works, we pick up the top 5 keyphrases in sequence order. The 5 leftmost keyphrases in the predicted sequence are selected as the result. 

\end{document}